\documentclass[11pt,a4paper]{article}
\usepackage[hyperref]{acl2020}
\usepackage{times}
\usepackage{latexsym}

\usepackage{microtype}

\usepackage{url}
\usepackage{graphicx}
\usepackage{booktabs}
\usepackage{xcolor}
\usepackage{comment}
\usepackage{amsmath}
\usepackage{amsfonts}
\usepackage{upgreek}
\usepackage{siunitx}
\usepackage{multirow}

\usepackage{natbib}
\usepackage{xspace}
\usepackage{soul}
\usepackage{adjustbox}
\usepackage{wrapfig}

\aclfinalcopy 

\title{One Size Does Not Fit All: Generating and Evaluating Variable Number of Keyphrases}

\author{Xingdi Yuan$^\dag$\thanks{\:\:\:\:These authors contributed equally. The order is determined by a fidget spinner.} \:\:\:\: Tong Wang$^{\dag}$\footnotemark[1] \:\:\:\: Rui Meng$^\ddag$\footnotemark[1] \:\:\:\: Khushboo Thaker$^{\ddag}$ \\ 
\textbf{Peter Brusilovsky$^{\ddag}$ \:\:\:\: Daqing He$^{\ddag}$ \:\:\:\: Adam Trischler$^{\dag}$}\\
$^\dag$Microsoft Research, Montr\'{e}al \\
$^\ddag$School of Computing and Information, University of Pittsburgh \\
\{eric.yuan, tong.wang\}@microsoft.com \:\:\:\: rui.meng@pitt.edu
}

\date{}

\usepackage{array}
\newcolumntype{C}[1]{>{\centering}m{#1}}

\newcounter{lintr}
\newcommand\TODO[1]{\textcolor{red}{\\TODO: #1\\}}

\newcommand{\lti}{{\arabic{lintr}\stepcounter{lintr}}}
\newcommand{\rlti}[1]{\arabic{lintr}\refstepcounter{lintr}\label{#1}}
\newcommand{\smx}{{\mathrm{softmax}}}
\newcommand{\sigm}{{\mathrm{sigmoid}}}

\newcommand{\GRU}{{\mathrm{GRU}}}
\newcommand\norm[1]{\left\lVert#1\right\rVert}
\newcommand{\sep}{{$\langle$\texttt{sep}$\rangle$}}
\newcommand{\eos}{{$<$\texttt{/s}$>$}}

\newcommand{\base}{\textbf{$\texttt{CatSeq}$}\xspace}  
\newcommand{\ours}{\textbf{$\texttt{CatSeqD}$}\xspace}  

\newcommand{\fk}{\textcolor{green1}{\textbf{F$_1$@$k$}}\xspace}
\newcommand{\fm}{\textcolor{blue1}{\textbf{F$_1$@$\mathcal{M}$}}\xspace}
\newcommand{\fo}{\textcolor{red1}{\textbf{F$_1$@$\mathcal{O}$}}\xspace}
 
\newcommand{\otoo}{\texttt{\textsc{One2One}}\xspace}
\newcommand{\otom}{\texttt{\textsc{One2Many}}\xspace}
\newcommand{\otos}{\texttt{\textsc{One2Seq}}\xspace}
\newcommand{\kpk}{\texttt{\textsc{KP20k}}\xspace}

\newcommand{\stackex}{\texttt{\textsc{StackEx}}\xspace}
\newcommand{\duc}{\texttt{\textsc{DUC}}\xspace}
\newcommand{\inspec}{\texttt{\textsc{Inspec}}\xspace}
\newcommand{\krapivin}{\texttt{\textsc{Krapivin}}\xspace}
\newcommand{\nus}{\texttt{\textsc{NUS}}\xspace}
\newcommand{\semeval}{\texttt{\textsc{SemEval}}\xspace}

\newcommand{\code}[1]{\texttt{#1}}
\newcommand{\cmd}[1]{\textbf{\small{\code{#1}}}}

\newcommand\numberthis{\addtocounter{equation}{1}\tag{\theequation}}

\definecolor{green1}{HTML}{257101}
\definecolor{blue1}{HTML}{2C05A5}
\definecolor{red1}{HTML}{A50520}

\begin{document}
\maketitle
\begin{abstract}
Different texts shall by nature correspond to different number of keyphrases. This desideratum is largely missing from existing neural keyphrase generation models. In this study, we address this problem from both modeling and evaluation perspectives.
 
We first propose a recurrent generative model that generates multiple keyphrases as delimiter-separated sequences. Generation diversity is further enhanced with two novel techniques by manipulating decoder hidden states.
In contrast to previous approaches, our model is capable of generating diverse keyphrases and controlling number of outputs.

We further propose two evaluation metrics tailored towards the variable-number generation. We also introduce a new dataset (\stackex) that expands beyond the only existing genre (i.e., academic writing) in keyphrase generation tasks. With both previous and new evaluation metrics, our model outperforms strong baselines on all datasets.
\end{abstract}

\section{Introduction}
\label{introduction}
Keyphrase generation is the task of automatically predicting keyphrases given a source text.
Desired keyphrases are often multi-word units that summarize the high-level meaning and highlight certain important topics or information of the source text. Consequently, models that can successfully perform this task should be capable of not only distilling high-level information from a document, but also locating specific, important snippets therein.

To make the problem even more challenging, a keyphrase may or may not be a substring of the source text (i.e., it may be \emph{present} or \emph{absent}).
Moreover, a given source text is usually associated with a \emph{set} of multiple keyphrases.
Thus, keyphrase generation is an instance of the set generation problem, where both the size of the set and the size (i.e., the number of tokens in a phrase) of each element can vary depending on the source.

Similar to summarization, keyphrase generation is often formulated as a sequence-to-sequence (Seq2Seq) generation task in most prior studies \citep{meng2017deep,chen2018kp_correlation,ye2018kp_semi,chen2018kp_title}.
Conditioned on a source text, Seq2Seq models generate phrases individually or as a longer sequence jointed by delimiting tokens.
Since standard Seq2Seq models generate only one sequence at a time, thus to generate multiple phrases, a common approach is to over-generate using beam search \citep{reddy1977speech} with a large beam width.
Models are then evaluated by taking a \textit{fixed} number of top predicted phrases (typically 5 or 10) and comparing them against the ground truth keyphrases.

\begin{table}[t!]
    \centering
    \scriptsize
    \begin{tabular}{r|c|c|c|c|c|c}
        \toprule
        \midrule
        Dataset & \#Train & \#Valid & \#Test & Mean & Var & \%Pre\\
        \midrule
        \midrule
        \text{\kpk} & $\approx$514k & $\approx$20k & $\approx$20k & 5.3 & 14.2 & 63.3\% \\
        \text{\inspec} & -- & 1500 & 500 & 9.6 & 22.4 & 78.5\% \\
        \text{\krapivin} & -- & 1844 & 460 & 5.2 & 6.6 & 56.2\% \\
        \text{\nus} & -- & - & 211 & 11.5 & 64.6 & 51.3\% \\
        \text{\semeval} & -- & 144 & 100 & 15.7 & 15.1 & 44.5\% \\
        \text{\stackex} & $\approx$298k & $\approx$16k & $\approx$16k & 2.7 & 1.4 & 57.5\% \\
        \midrule
        \bottomrule
    \end{tabular}
    \caption{Statistics of various datasets. Mean and Var indicate the mean and variance of target phrase numbers, \%Pre denotes percentage of present keyphrases.}
    \label{tab:kp_data_stats}
\end{table}

Though this approach has achieved good empirical results, we argue that it suffers from two major limitations.
Firstly, models that use beam search to generate multiple keyphrases generally lack the ability to determine the dynamic number of keyphrases needed for different source texts.
Meanwhile, the parallelism in beam search also fails to model the inter-relation among the generated phrases, which can often result in diminished diversity in the output.
Although certain existing models take output diversity into consideration during training \citep{chen2018kp_correlation,ye2018kp_semi},
the effort is significantly undermined during decoding due to the reliance on over-generation and phrase ranking with beam search.

Secondly, the current evaluation setup is rather problematic, since existing studies attempt to match a fixed number of outputs against a variable number of ground truth keyphrases.
Empirically, the number of keyphrases can vary drastically for different source texts, depending on a plethora of factors including the length or genre of the text, the granularity of keyphrase annotation, etc. For the several commonly used keyphrase generation datasets, for example, the average number of keyphrases per data point can range from 5.3 to 15.7, with variances sometimes as large as 64.6 (Table~\ref{tab:kp_data_stats}).
Therefore, using an arbitrary, fixed number $k$ to evaluate entire datasets is not appropriate.
In fact, under this evaluation setup, the F1 score for the oracle model on the \kpk dataset is 0.858 for $k=5$ and 0.626 for $k=10$, which apparently poses serious normalization issues as evaluation metrics.

To overcome these problems, we propose novel decoding strategies and evaluation metrics for the keyphrase generation task. 
The main contributions of this work are as follows:
\begin{enumerate}
    \item We propose a Seq2Seq based keyphrase generation model capable of generating diverse keyphrases and controlling number of outputs. 
    \item We propose \textbf{new metrics} based on commonly used $F_{1}$ score under the hypothesis of \textit{variable-size} outputs from models, which results in improved empirical characteristics over previous metrics based on a fixed $k$.
    \item An additional contribution of our study is the introduction of a \textbf{new dataset} for keyphrase generation: \stackex.
\end{enumerate}
With its marked difference in genre, we expect the dataset to bring added heterogeneity to keyphrase generation evaluation.

\section{Related Work}
\label{sec:related_work}

\subsection{Keyphrase Extraction and Generation}
Traditional keyphrase extraction has been studied extensively in past decades. In most existing literature, keyphrase extraction has been formulated as a two-step process. 
First, lexical features such as part-of-speech tags are used to determine a list of phrase candidates by heuristic methods~\citep{witten1999kea,liu2011gap,wang2016ptr,yang2017semisupervisedqa}. 
Second, a ranking algorithm is adopted to rank the candidate list and the top ranked candidates are selected as keyphrases. 
A wide variety of methods were applied for ranking, such as bagged decision trees~\citep{medelyan2009human_competitive,lopez2010humb}, Multi-Layer Perceptron, Support Vector Machine~\citep{lopez2010humb} and PageRank~\citep{Mihalcea2004textrank,le2016unsupervised,wan2008neighborhood_knowledge}.
Recently, \citet{zhang2016twitter,luan2017tagging,gollapalli2017expert_knowledge} used sequence labeling models to extract keyphrases from text;
\citet{subramanian2017kp} used Pointer Networks to point to the start and end positions of keyphrases in a source text;
\citet{sun2019divgraphpointer} leveraged graph neural networks to extract keyphrases.

The main drawback of keyphrase extraction is that sometimes keyphrases are absent from the source text, thus an extractive model will fail predicting those keyphrases. 
\citet{meng2017deep} first proposed the CopyRNN, a neural model that both generates words from vocabulary and points to words from the source text. 
Based on the CopyRNN architecture, \citet{chen2018kp_correlation, zhao2019incorporating} leveraged attention to help reducing duplication and improving coverage.
\citet{ye2018kp_semi} proposed semi-supervised methods by leveraging both labeled and unlabeled data for training. 
\citet{chen2018kp_title,ye2018kp_semi} proposed to use structure information (e.g., title of source text) to improve keyphrase generation performance.
\citet{chan2019neural} introduced RL to the keyphrase generation task.
\citet{chen2019integrated} retrieved similar documents from training data to help producing more accurate keyphrases.

\subsection{Sequence to Sequence Generation}
Sequence to Sequence (Seq2Seq) learning was first introduced by \citet{sutskever2014seq2seq}; together with the soft attention mechanism of~\citep{bahdanau2014attntion}, it has been widely used in natural language generation tasks. \citet{gulcehre2016pointersoftmax,Gu2016copy} used a mixture of generation and pointing to overcome the problem of large vocabulary size. \citet{paulus2017summarization,zhou2017summarization} applied Seq2Seq models on summary generation tasks, while \citet{du2017qgen,yuan2017qgen} generated questions conditioned on documents and answers from machine comprehension datasets. Seq2Seq was also applied on neural sentence simplification~\citep{zhang2017simplification} and paraphrase generation tasks~\citep{xu2018dpage}.

\section{Model Architecture}
\label{sec:model}
Given a piece of source text, our objective is to generate a variable number of multi-word phrases. To this end, we opt for the sequence-to-sequence (Seq2Seq) \citep{sutskever2014seq2seq} framework as the basis of our model, combined with attention and pointer softmax mechanisms in the decoder. 

Since each data example contains one source text sequence and multiple target phrase sequences (dubbed \otom, and each sequence can be of multi-word), two paradigms can be adopted for training Seq2Seq models. The first one~\citep{meng2017deep} is to divide each \otom data example into multiple \otoo examples, and the resulting models (e.g., CopyRNN) can generate one phrase at once and must rely on beam search technique to produce more unique phrases. 

\begin{figure*}[t]
    \centering
    \includegraphics[width=0.85\textwidth]{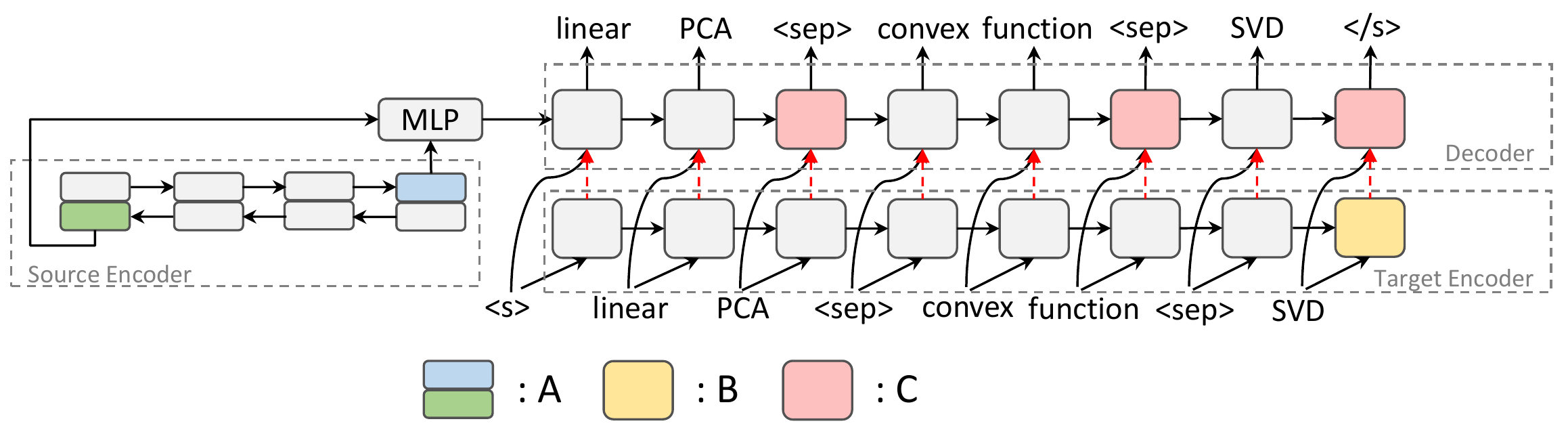}
    \caption{The architecture of the proposed model for improving keyphrase diversity. $A$ represents last states of a bi-directional source encoder; $B$ represents the last state of target encoder; $C$ indicates decoder states where target tokens are either delimiters or end-of-sentence tokens. During orthogonal regularization, all $C$ states are used; during target encoder training, we maximize mutual information between states $A$ with $B$. Red dash arrow indicates a detached path, i.e., no back-propagation through such path.}
    \label{fig:model_structure}
\end{figure*}

To enable models to generate multiple phrases and control the number to output, we propose the second training paradigm \otos, in which we concatenate multiple phrases into a single sequence with a delimiter \sep, and this concatenated sequence is then used as the target for sequence generation during training. An overview of the model's structure is shown in Figure~\ref{fig:model_structure}.\footnote{We release the code, datasets and model outputs for reproducing our results in \url{https://github.com/memray/OpenNMT-kpg-release}.}

\subsection*{Notations}
In the following subsections, we use $w$ to denote input text tokens, $x$ to denote token
embeddings, $h$ to denote hidden states, and $y$ to denote output text tokens. 
Superscripts denote time-steps in a sequence, and
subscripts $e$ and $d$ indicate whether a variable resides in the encoder or the
decoder of the model, respectively. 
The absence of a superscript indicates multiplicity in the time dimension.
$L$ refers to a linear transformation and $L^{f}$ refers to it followed by a non-linear activation function $f$. 
Angled brackets, $\langle\rangle$, denote concatenation.

\subsection{Sequence to Sequence Generation}
We develop our model based on the standard Seq2Seq \citep{sutskever2014seq2seq} model with attention mechanism \citep{bahdanau2014attntion} and pointer softmax \citep{gulcehre2016pointersoftmax}.
Due to space limit, we describe this basic Seq2Seq model in Appendix~\ref{appd:basic_model}.

\subsection{Mechanisms for Diverse Generation}

There are usually multiple keyphrases for a given source text because each keyphrase represents certain aspects of the text. 
Therefore keyphrase diversity is desired for the keyphrase generation.
Most previous keyphrase generation models generate multiple phrases by over-generation, which is highly prone to generate similar phrases due to the nature of beam search. 
Given our objective to generate variable numbers of keyphrases, we need to adopt new strategies for achieving better diversity in the output.

Recall that we represent variable numbers of keyphrases as delimiter-separated sequences. One particular
issue we observed during error analysis is that the model tends to produce identical tokens 
following the delimiter token. For example, suppose 
a target sequence contains $n$ delimiter tokens at time-steps $t_1,\dots,t_n$. During training, the model is rewarded for generating the same delimiter
token at these time-steps, which presumably introduces much homogeneity in the corresponding decoder 
states $h_d^{t_1},\dots,h_d^{t_n}$.  
When these states are subsequently used as inputs at the time-steps
immediately following the delimiter, the decoder naturally produces highly
similar distributions over the following tokens, resulting in identical tokens being decoded. 
To alleviate this problem, we propose two plug-in components for the sequential generation model.

\subsubsection{Semantic Coverage}
\label{sec:target_enc}

We propose a mechanism called \textit{semantic coverage} that focuses on the semantic representations of generated phrases. 
Specifically, we introduce another uni-directional recurrent model $\GRU_\mathrm{SC}$ (dubbed \textit{target encoder}) which encodes decoder-generated tokens $y^\tau$, where $\tau \in [0, t)$, into hidden states $h_\mathrm{SC}^t$.
This state is then taken as an extra input to the decoder GRU, modifying equation of the decoder GRU to:
\begin{equation}
h_d^t=\GRU_d(\langle x_d^{t},h_\mathrm{SC}^t\rangle,h_d^{t-1}).
\end{equation}

If the target encoder were to be updated with the training signal from generation (i.e., backpropagating error from the decoder GRU to the target encoder),
the resulting decoder is essentially a 2-layer GRU with residual connections.
Instead, inspired by previous representation learning works \citep{logeswaran2017representations,vandenoord2018cpc,hjelm2018learning}, we train the target encoder in an self-supervised fashion (Figure~\ref{fig:model_structure}).
Specifically, due to the autoregressive nature of the RNN-based decoder, we follow Contrastive Predictive Coding (CPC) \citep{vandenoord2018cpc}, where a Noise-Contrastive Estimation(NCE) loss is used to maximize a lower bound on mutual information.
That is, we extract target encoder's final hidden state vector $h_\mathrm{SC}^M$, where $M$ is the length of target sequence, and use it as a general representation of the target phrases.
We train by maximizing the mutual information between these phrase representations and the final state of the source encoder $h_e^T$ as follows. 
For each phrase representation vector $h_\mathrm{SC}^M$, we take the encodings $H_e^T = \{h_{e,1}^T,\dots,h_{e,N}^T\}$ of $N$ different source texts,
where $h_{e,true}^T$ is the encoder representation for the current source text,
and the remaining $N-1$ are negative samples (sampled at random) from the training data.
The target encoder is trained to minimize the classification loss:
\begin{equation}
\begin{aligned}
\label{eqn:target_encoder_loss}
\mathcal{L_{\mathrm{SC}}}&=-\mathrm{log}\frac{g(h_{e,true}^T, h_\mathrm{SC}^M)}{\sum_{i \in [1,N]}{g(h_{e,i}^T, h_\mathrm{SC}^M)}}, \\
g(h_a, h_b) &= \mathrm{exp}(h_a^\top Bh_b)\\
\end{aligned}
\end{equation}
where $B$ is bi-linear transformation.

The motivation here is to constrain the overall representation of generated keyphrase to be semantically
close to the overall meaning of the source text. 
With such representations as input to the decoder, the semantic coverage mechanism can potentially help to provide useful keyphrase information and guide generation.

\subsubsection{Orthogonal Regularization}
\label{sec:orthogonal_reg}

We also propose orthogonal regularization, which explicitly encourages the delimiter-generating decoder states to be
different from each other. 
This is inspired by \citet{bousmalis2016domainsep}, who use orthogonal regularization to encourage representations across domains to be as distinct as possible.
Specifically, we stack the decoder hidden states corresponding to delimiters together to form matrix $H=\langle h_d^{t_1},\dots,h_d^{t_n}\rangle$ and use the following equation as the orthogonal regularization loss:
\begin{equation}
\mathcal{L_{\mathrm{OR}}}=\norm{H^\top H \odot (\textbf{1}-I_n)}_2,
\end{equation}
where $H^\top$ is the matrix transpose of $H$, $I_n$ is the identity matrix of rank $n$, $\odot$ indicates element wise multiplication, $\norm{M}_2$ indicates $L^2$ norm of each element in a matrix $M$. 
This loss function prefers orthogonality among
the hidden states $h_d^{t_1},\dots,h_d^{t_n}$ and thus improves diversity in the tokens following the delimiters.

\subsubsection{  Training Loss}
We adopt the widely used negative log-likelihood loss in our sequence generation model, denoted as $\mathcal{L_{\mathrm{NLL}}}$. The overall loss we use for optimization is:
\begin{equation}
\label{eqn:entire_loss}
\mathcal{L}=\mathcal{L_{\mathrm{NLL}}} +
\lambda_\mathrm{OR} \cdot \mathcal{L_{\mathrm{OR}}} +
\lambda_\mathrm{SC} \cdot \mathcal{L_{\mathrm{SC}}},
\end{equation}
where $\lambda_\mathrm{OR}$ and $\lambda_\mathrm{SC}$ are hyper-parameters.

\subsection{Decoding Strategies}
\label{subsec:decoding_strategies}
According to different task requirements, various decoding methods can be applied to generate the target sequence $y$. Prior studies
\citet{meng2017deep,yang2017semisupervisedqa}
focus more on generating excessive number of phrases by leveraging beam search to proliferate the output phrases. In contrast, models trained under \otos paradigm are capable of determining the proper number of phrases to output. In light of previous research in psychology~\citep{van1993self,forster1976terminating}, we name these two decoding/search strategies as Exhaustive Decoding and Self-terminating Decoding, respectively, due to their resemblance to the way humans behave in serial memory tasks. Simply speaking, the major difference lies in whether a model is capable of controlling the number of phrases to output. We describe the detailed decoding strategies used in this study as follows:


\subsubsection{  Exhaustive Decoding}
\label{subsec:fixed_size}
As traditional keyphrase tasks evaluate models with a fixed number of top-ranked predictions (say F-score @5 and @10), existing keyphrase generation studies have to over-generate phrases by means of beam search (commonly with a large beam size, e.g., 150 and 200 in \citep{chen2018kp_title,meng2017deep}, respectively), a heuristic search algorithm that returns $K$ approximate optimal sequences. For the \otoo setting, each returned sequence is a unique phrase itself. But for \otos, each produced sequence contains several phrases and additional processes~ \citep{ye2018kp_semi} are needed to obtain the final unique (ordered) phrase list.

It is worth noting that the time complexity of beam search is $O(Bm)$, where $B$ is the beam width, and $m$ is the maximum length of generated sequences. Therefore the exhaustive decoding is generally very computationally expensive, especially for \otos setting where $m$ is much larger than in \otoo. It is also wasteful as we observe that less than 5\% of phrases generated by \otos models are unique.

\subsubsection{Self-terminating Decoding}
\label{subsec:variable_size}
An innate characteristic of keyphrase tasks is that the number of keyphrases varies depending on the document and dataset genre, therefore dynamically outputting a variable number of phrases is a desirable property for keyphrase generation models \footnote{Note this is fundamentally different from other NLG tasks. In specific, the number of keyphrases is variable, the length of each keyphrase is also variable.}. 
Since our model is trained to generate a variable number of phrases as a single sequence joined by delimiters, we can obtain multiple phrases by simply decoding a single sequence for each given source text. The resulting model thus implicitly performs the additional task of dynamically estimating the proper size of the target phrase set: once the model believes that an adequate number of phrases have been generated, it outputs a special token \eos~to terminate the decoding process. 

One notable attribute of the self-terminating decoding strategy is that, by generating a set of phrases in a single sequence, the model conditions its current generation on all previously generated phrases. Compared to the exhaustive strategy (i.e., phrases being generated independently by beam search in parallel), our model can model the dependency among its output in a more explicit fashion. Additionally, since multiple phrases are decoded as a single sequence, decoding can be performed more efficiently than exhaustive decoding by conducting greedy search or beam search on only the top-scored sequence. 

\section{Evaluating Keyphrase Generation}
\label{sec:evaluation_method}

Formally, given a source text, suppose that a model predicts a list of unique keyphrases $\mathcal{\hat Y}=(\hat y_1,\dots,\hat y_m)$ ordered by the quality of the predictions $\hat y_i$, and that the ground truth keyphrases for the given source text is the oracle set $\mathcal{Y}$. When only the top $k$ predictions $\mathcal{\hat Y}_{:k}=(\hat y_1,\dots,\hat y_{\min(k,m)})$ are used for evaluation, \textit{precision}, \textit{recall}, and \textit{F$_1$} score are consequently conditioned on $k$ and defined as:
\begin{equation}
\begin{aligned}
    P@k &= \frac{|\mathcal{\hat Y}_{:k}\cap\mathcal{Y}|}{|\mathcal{\hat Y}_{:k}|}, \:\:\:\:\:\:\:\:
    R@k = \frac{|\mathcal{\hat Y}_{:k}\cap\mathcal{Y}|}{|\mathcal{Y}|},   \\
    \fk &= \frac{2 * P@k * R@k}{P@k + R@k}.
\end{aligned}
\end{equation}

As discussed in Section~\ref{introduction}, the number of generated keyphrases used for evaluation can have a critical impact on the quality of the resulting evaluation metrics. Here we compare three choices of $k$ and the implications on keyphrase evaluation for each choice:\\
$\bullet$ \fk: where $k$ is a pre-defined constant (usually 5 or 10). Due to the high variance of the number\ of ground truth keyphrases, it is often that $|\mathcal{\hat Y}_{:k}|\leq k<|\mathcal{Y}|$, and thus $R@k$ --- and in turn $F_1@k$ --- of an oracle model can be smaller than $1$. This undesirable property is unfortunately prevalent in the evaluation metrics adopted by all existing keyphrase generation studies to our knowledge.

\begin{table*}[ht!]
  \centering
  \scriptsize
  \begin{tabular}{c|C{0.3cm}C{0.3cm}C{0.5cm}|C{0.3cm}C{0.3cm}C{0.5cm}|C{0.3cm}C{0.3cm}C{0.5cm}|C{0.3cm}C{0.3cm}C{0.5cm}|C{0.3cm}C{0.3cm}c}
    \toprule
    \midrule
    & \multicolumn{3}{c|}{\textbf{Kp20K}}
    & \multicolumn{3}{c|}{\textbf{Inspec}}
    & \multicolumn{3}{c|}{\textbf{Krapivin}}
    & \multicolumn{3}{c|}{\textbf{NUS}}
    & \multicolumn{3}{c}{\textbf{SemEval}}
    \\
    \midrule
    Model & \tiny{\textbf{F$_1$@5}}& \tiny{\textbf{F$_1$@10}} & \tiny{\fo} & \tiny{\textbf{F$_1$@5}}& \tiny{\textbf{F$_1$@10}} & \tiny{\fo} & \tiny{\textbf{F$_1$@5}}& \tiny{\textbf{F$_1$@10}} & \tiny{\fo} & \tiny{\textbf{F$_1$@5}}& \tiny{\textbf{F$_1$@10}} & \tiny{\fo} & \tiny{\textbf{F$_1$@5}}& \tiny{\textbf{F$_1$@10}} & \tiny{\fo}\\  
    \midrule
    \multicolumn{16}{c}{\textbf{Abstractive Neural}} \\
    \midrule
    \textbf{CopyRNN}~(\citeauthor{meng2017deep}) & 32.8 & 25.5 & -- & \underline{29.2} & \underline{33.6} & --    & 30.2 & 25.2 & --    & 34.2 & 31.7 & --    & 29.1 & 29.6 & --  \\  
    \textbf{CopyRNN*} 
    & 31.7 & 27.3 & 33.5 & 24.4 & 28.9 & 29.0 & 30.5 & 26.6
    & 32.5 & \textbf{37.6} & \underline{35.2} & \textbf{40.6} & 31.8 & 31.8 & \underline{31.7}  \\  
    \textbf{CorrRNN}~(\citeauthor{chen2018kp_correlation}) & - & - & - & - & - & - & 31.8 & 27.8 & - & 35.8 & 33.0 & - & \underline{32.0} & \underline{32.0} & -  \\ 
    \tiny{\textbf{ParaNetT +CoAtt}~(\citeauthor{zhao2019incorporating})} 
    & \textbf{36.0} & 28.9 & - 
    & \textbf{29.6} & \textbf{35.7} & - 
    & \underline{32.9} & 28.2 & - 
    & 36.0 & 35.0 & - 
    & 31.1 & 31.2 & -  \\ 
     
    \textbf{catSeqTG-2RF1$^\dag$}~(\citeauthor{chan2019neural}) 
    & 32.1 & - & \underline{35.7}
    & 25.3 & - & 28.0 
    & 30.0 & - & \underline{34.8}
    & \underline{37.5} & - & 25.5 
    & 28.7 & - & 29.8  \\ 
    \textbf{KG-KE-KR-M$^\dag$}~(\citeauthor{chen2019integrated}) 
    & 31.7 & 28.2 & \textbf{38.8}  
    & 25.7 & 28.4 & \underline{31.4}
    & 27.2 & 25.0 & 31.7
    & 28.9 & 28.6 & 38.4
    & 20.2 & 22.3 & 30.3  \\  
    
    \text{\base (Ours)} 
    & 31.4 & 27.3 & 31.9 
    & 29.0 & 30.0 & 30.7 
    & 30.7 & 27.4 & 32.4 
    & 35.9 & 34.9 & \underline{38.3} 
    & 30.2 & 30.6 & 31.0  \\  
    \text{\ours (Ours)} 
    & 34.8 & \underline{29.8} & \underline{35.7} 
    & 27.6 & 33.3 & \textbf{33.1} 
    & 32.5 & \underline{28.5} & \textbf{37.1} 
    & 37.4 & \textbf{36.6} & \textbf{40.6} 
    & \textbf{32.7} & \textbf{35.2} & \textbf{35.7}  \\  
    \midrule
    \multicolumn{16}{c}{\textbf{Extractive IR}} \\
    \midrule
    \textbf{TfIdf}~(\citeauthor{hasan2010conundrums}) & 7.2 & 9.4 & 6.3 & 16.0 & 24.4 & 20.8 & 6.7 & 9.3 & 6.8 & 11.2 & 14.0 & 12.2 & 8.8 & 14.7 &  11.3 \\
    \textbf{TextRank}~(\citeauthor{Mihalcea2004textrank}) & 18.1 & 15.1 & 18.4 & 28.6 & 33.9 & 33.5 & 18.5 & 16.0 & 21.1 & 23.0 & 21.6 & 23.8 & 21.7 & 22.6 & 22.9 \\ 
    \textbf{KEA}~(\citeauthor{witten1999kea}) & 4.6 & 4.4 & 5.1 & 2.2 & 2.2 & 2.2  & 1.8 & 1.7 & 1.7 & 7.3 & 7.1 & 8.1 & 6.8 & 6.5 & 6.6 \\ 
    \textbf{Maui}~(\citeauthor{medelyan2009human_competitive}) & 0.5 & 0.5 & 0.4 & 3.5 & 4.6 & 3.9 & 0.5 & 0.7 & 0.6 & 0.4 & 0.6 & 0.6 & 1.1 & 1.4 & 1.1 \\ 
    \midrule
    \multicolumn{16}{c}{\textbf{Extractive Neural}} \\
    \midrule
    \textbf{DivGraphPointer}~(\citeauthor{sun2019divgraphpointer}) & 36.8 & 29.2 & - & 38.6 & 41.7 & - & 46.0 & 40.2 & - & 40.1 & 38.9 & - & 36.3 & 29.7 & -  \\ 
    \midrule
    \multicolumn{16}{c}{\textbf{w/ Additional Data}} \\
    \midrule
    \textbf{Semi-Multi}~(\citeauthor{ye2018kp_semi}) & 32.8 & 26.4 & - & 32.8 & 31.8 & - & 32.3 & 25.4 & - & 36.5 & 32.6 & - & 31.9 & 31.2 & -  \\  
    \textbf{TG-Net}~(\citeauthor{chen2019guided}) & 37.2 & 31.5 & - & 31.5 & 38.1 & - & 34.9 & 29.5 & - & 40.6 & 37.0 & - & 31.8 & 32.2 & - \\
    \bottomrule
  \end{tabular}
  \caption{Performance of present keyphrase prediction on scientific publications datasets. Best/second-best performing score in each column is highlighted with bold/underline. We also list results from literature where models that are not directly comparable (i.e., models leverage additional data and pure extractive models). Note model names with $^\dag$ represent its \fo is computed by us using existing works' released keyphrase predictions.\footnotemark}
  \label{tab:transfer}
\end{table*}


A simple remedy is to set $k$ as a variable number which is specific to each data example. Here we define two new metrics:\\
$\bullet$ \fo: \textbf{$\mathcal{O}$} denotes the number of oracle (ground truth) keyphrases. In this case, $k=|\mathcal{Y}|$, which means for each data example, the number of predicted phrases taken for evaluation is the same as the number of ground truth keyphrases.\\
$\bullet$ \fm: \textbf{$\mathcal{M}$} denotes the number of predicted keyphrases. In this case, $k=|\mathcal{\hat Y}|$ and we simply take all the predicted phrases for evaluation without truncation.

By simply extending the \textit{constant} number $k$ to different variables accordingly, both \fo and \fm are capable of reflecting the nature of \textit{variable} number of phrases for each document, and a model can achieve the maximum $F_1$ score of $1.0$ if and only if it predicts the exact same phrases as the ground truth. Another merit of \fo is that it is independent from model outputs, therefore we can use it to compare existing models.



\footnotetext{We acknowledge that \fo scores of \citet{chan2019neural} and \citet{chen2019integrated} might be not completely comparable with ours. This is due to additional post-processing and filtering methods might have been applied in different work. We elaborate the data pre-processing and evaluation protocols used in this work in Appendix~\ref{appd:evaluate_detail}.}

\section{Datasets and Experiments}
\label{sec:dataset_exp}
In this section, we report our experiment results on multiple datasets and compare with existing models.
We use \base to refer to the delimiter-concatenated sequence-to-sequences model described in Section~\ref{sec:model}; \ours refers to the model augmented with orthogonal regularization and semantic coverage mechanism.

To construct target sequences for training \base and \ours, ground truth keyphrases are sorted by their order of first occurrence in the source text. 
Keyphrases that do not appear in the source text are appended to the end. 
This order may guide the attention mechanism to attend to source positions in a smoother way. Implementation details can be found in Appendix~\ref{appd:implement_detail}. As for the pre-processing and evaluation, we follow the same steps as in~\cite{meng2017deep}. More details are provide in Appendix~\ref{appd:evaluate_detail} for reproducing our results.


We include a set of existing models \citep{meng2017deep, chen2018kp_correlation, chan2019neural, zhao2019incorporating, chen2019integrated} as baselines, they all share same behavior of abstractive keyphrase generation with our proposed model. 
Specially for computing existing model's scores with our proposed new metrics (\fo and \fm), we implemented our own version of CopyRNN \citep{meng2017deep} based on their open sourced code, denoted as CopyRNN*. We also report the scores of models from \citeauthor{chan2019neural} and \citeauthor{chen2019integrated} based on their publicly released outputs.

We also include a set of models that use similar strategies but can not directly compare with.
This includes four non-neural extractive models: TfIdf \citep{hasan2010conundrums}, TextRank \citep{Mihalcea2004textrank}, KEA \citep{witten1999kea}, and Maui \citep{medelyan2009human_competitive}; 
one neural extractive model \citep{sun2019divgraphpointer};
and two neural models that use additional data (e.g., title) \citep{ye2018kp_semi, chen2019guided}.


In Section~\ref{sec:f1_at_m}, we apply the self-terminating decoding strategy. Since no existing model supports such decoding strategy, we only report results from our proposed models. 
They can be used for comparison in future studies.


\subsection{Experiments on Scientific Publications}
\label{subsec:scientific}



Our first dataset consists of a collection of scientific publication datasets, namely \kpk, \inspec, \krapivin, \nus, and \semeval, that have been widely used in existing literature \citep{meng2017deep,chen2018kp_correlation,ye2018kp_semi,chen2018kp_title,chan2019neural,zhao2019incorporating,chen2019integrated,sun2019divgraphpointer}.
\kpk, for example, was introduced by \citet{meng2017deep} and comprises more than half a million scientific publications. 
For each article, the abstract and title are used as the source text while the author keywords are used as target.
The other four datasets contain much fewer articles, and thus used to test transferability of our model.


\begin{table}[t!]
    \centering
    \scriptsize
    \begin{tabular}{r|c|c|c|c|c}
        \toprule
        \midrule
        & \multicolumn{3}{c|}{Present} & \multicolumn{2}{c}{Absent} \\
        \midrule
        Model & \textbf{F$_1$@5} & \textbf{F$_1$@10} & \fo & \textbf{R@10} & \textbf{R@50} \\
        \midrule
        \midrule
        \text{TfIdf} & 8.0 & 8.9 & 5.2 & - & - \\
        \text{TextRank} & 12.1 & 10.1 & 11.6  & - & -\\
        \text{KEA} & 4.9 & 4.8 & 5.3  & - & -\\
        \text{Maui} & 35.8 & 23.3 & 51.8 & - & - \\
        \midrule
        \midrule
        \text{CopyRNN*} & 44.2 & 30.3 & \textbf{66.2} & \underline{48.8} & \textbf{66.0} \\
        \text{\base} & \underline{48.3} & \textbf{45.5} & 63.5 & 40.7 & 42.2 \\
        \text{\ours} & \textbf{48.7} & \underline{43.9} & \underline{65.6} & \textbf{54.8} & \underline{65.7} \\
        \midrule
        \bottomrule
    \end{tabular}
    \caption{Model performance on \stackex dataset.}
    \label{tab:stack_exchange}
\end{table}

We report our model's performance on the \emph{present}-keyphrase portion of the \kpk dataset in Table~\ref{tab:transfer}.\footnote{We show experiment results on absent data in Appendix~\ref{appd:kp20k_absent}.} To compare with previous works, we provide compute $F_{1}@5$ and $F_{1}@10$ scores. The new proposed \fo metric indicates consistent ranking with $F_{1}@5/10$ for most cases. Due to its target number sensitivity, we find that its value is closer to $F_{1}@5$ for \kpk and \krapivin where average target keyphrases is less and closer to $F_{1}@10$ for the other three datasets.

From the result we can see that our \ours outperform \textit{existing abstractive models} on most of the datasets.
Our implemented CopyRNN* achieves better or comparable performance against the original model, and on NUS and SemEval the advantage is more salient. 

As for the proposed models, both \base and \ours yield comparable results to CopyRNN, indicating that \otos paradigm can work well as an alternative option for the keyphrase generation task. \ours outperforms \base on all metrics, suggesting the semantic coverage and orthogonal regularization help the model to generate higher quality keyphrases and achieve better generalizability. To our surprise, on the metric F$_1$@10 for \kpk and \krapivin (average number of keyphrases is only 5), where high-recall models like CopyRNN are more favored, \ours is still able to outperform \otoo baselines, indicating that the proposed mechanisms for diverse generation are effective.


\begin{table}[t!]
    \centering
    \scriptsize
    \begin{tabular}{r|c|c|c|c}
        \toprule
        \midrule
        & \multicolumn{2}{c|}{\kpk} & \multicolumn{2}{c}{\stackex} \\
        \midrule
        Model & \fo & \fm & \fo & \fm \\
        \midrule 
        \midrule
        \multicolumn{5}{c}{Greedy Search} \\
        \midrule
        \text{\base} & \underline{33.1} & 32.4 & \underline{59.2} & 56.3 \\
        \text{\ours} & \textbf{33.4} & \textbf{33.9} & \textbf{59.6} & \textbf{59.3} \\
        \midrule 
        \midrule
        \multicolumn{5}{c}{Top Ranked Sequence in Beam Search} \\
        \midrule
        \text{\base} & 24.3 & 25.1 & 52.4 & 52.7 \\
        \text{\ours} & 31.9 & \underline{33.4} & 56.5 & \underline{57.0} \\
        \midrule
        \bottomrule
    \end{tabular}
    \caption{\fo and \fm when generating variable number of keyphrases (self-terminating decoding).}
    \label{tab:f1_at_m}
\end{table}

\subsection{Experiments on The \stackex Dataset}
\label{sec:stack_exchange}

Inspired by the StackLite tag recommendation task on Kaggle, we build a new benchmark based on the public StackExchange data\footnote{https://archive.org/details/stackexchange, we choose 19 computer science related topics from Oct. 2017 dump.}. 
We use questions with titles as source, and user-assigned tags as target keyphrases.
We provide details regarding our data collection in Appendix~\ref{appd:stackex_collection}.

Since oftentimes the questions on StackExchange contain less information than in scientific publications, there are fewer keyphrases per data point in \stackex (statistics are shown in Table~\ref{tab:kp_data_stats}). 
Furthermore, StackExchange uses a tag recommendation system that suggests topic-relevant tags to users while submitting questions; therefore, we are more likely to see general terminology such as \cmd{Linux} and \cmd{Java}\footnote{One example is shown in Appendix~\ref{appd:cherry_pick}.}. This characteristic challenges models with respect to their ability to distill major topics of a question rather than selecting specific snippets from the text.



We report our models' performance on \stackex in Table~\ref{tab:stack_exchange}. Results show \ours performs the best in general; on the \emph{absent}-keyphrase generation tasks, it outperforms \base by a large margin.

\subsection{Generating Variable Number Keyphrases}
\label{sec:f1_at_m}

One key advantage of our proposed model is the capability of predicting the number of keyphrases conditioned on the given source text.
We thus conduct a set of experiments on \kpk and \stackex present keyphrase generation tasks, as shown in Table~\ref{tab:f1_at_m}, to study such behavior.
We adopt the self-terminating decoding strategy (Section~\ref{subsec:decoding_strategies}), and use both \fo and \fm (Section~\ref{sec:evaluation_method}) to evaluate.

In these experiments, we use beam search as in most Natural Language Generation (NLG) tasks, i.e., only use the top ranked prediction sequence as output. 
We compare the results with greedy search.
Since no existing model is capable of generating variable number of keyphrases, in this subsection we only report performance on such setting from \base and \ours.

From Table~\ref{tab:f1_at_m} we observe that in the variable number generation setting, greedy search outperforms beam search consistently. 
This may because beam search tends to generate short and similar sequences.
We can also see the resulting \fo scores are generally lower than results reported in previous subsections, this suggests an over-generation decoding strategy may still benefit from achieving higher recall.

\section{Analysis and Discussion}
\label{sec:analysis}

\subsection{Ablation Study}
\label{sec:ablation}

\begin{table}[t!]
    \centering
    \scriptsize
    \begin{tabular}{l|c|c|c|c|c}
        \toprule
        \midrule
        \textbf{Model} & \textbf{\kpk} &  \textbf{Inspec} & \textbf{Krapivin} & \textbf{NUS} & \textbf{SemEval} \\
        \midrule
        \midrule
        \base                   & 31.9 & 30.7 & 32.3 & 38.3 & 31.0 \\
        \:\:\:\:+ Orth. Reg.      & 31.1 & 29.3 & 31.0 & 36.5 & 29.5 \\
        \:\:\:\:+ Sem. Cov.       & \underline{32.9} & \underline{32.1} & \underline{34.5} & \underline{40.2} & \underline{32.9} \\
        \ours                   & \textbf{35.7} & \textbf{33.1} & \textbf{37.1} & \textbf{40.6} & \textbf{35.7} \\
        \midrule
        \bottomrule
    \end{tabular}
    \caption{Ablation study with \fo scores on five scientific publication datasets.}
    \label{tab:ablation}
\end{table}

We conduct an ablation experiment to study the effects of orthogonal regularization and semantic coverage mechanism on \base.
As shown in Table~\ref{tab:ablation}, semantic coverage provides significant boost to \base's performance on all datasets. 
Orthogonal regularization hurts performance when is solely applied to \base model.
Interestingly, when both components are enabled (\ours), the model outperforms \base by a noticeable margin on all datasets, this suggests the two components help keyphrase generation in a synergistic way.
One future direction is to apply orthogonal regularization directly on target encoder, since the regularizer can potentially diversify target representations at phrase level, which may further encourage diverse keyphrase generation in decoder.

\subsection{Visualizing Diversified Generation}
\label{subsection:tsne}
To verify our assumption that target encoding and orthogonal regularization help to boost the diversity of generated sequences, we use two metrics, one quantitative and one qualitative, to measure diversity of generation. 

\begin{figure}[t!]
    \centering
    \includegraphics[width=0.45\textwidth]{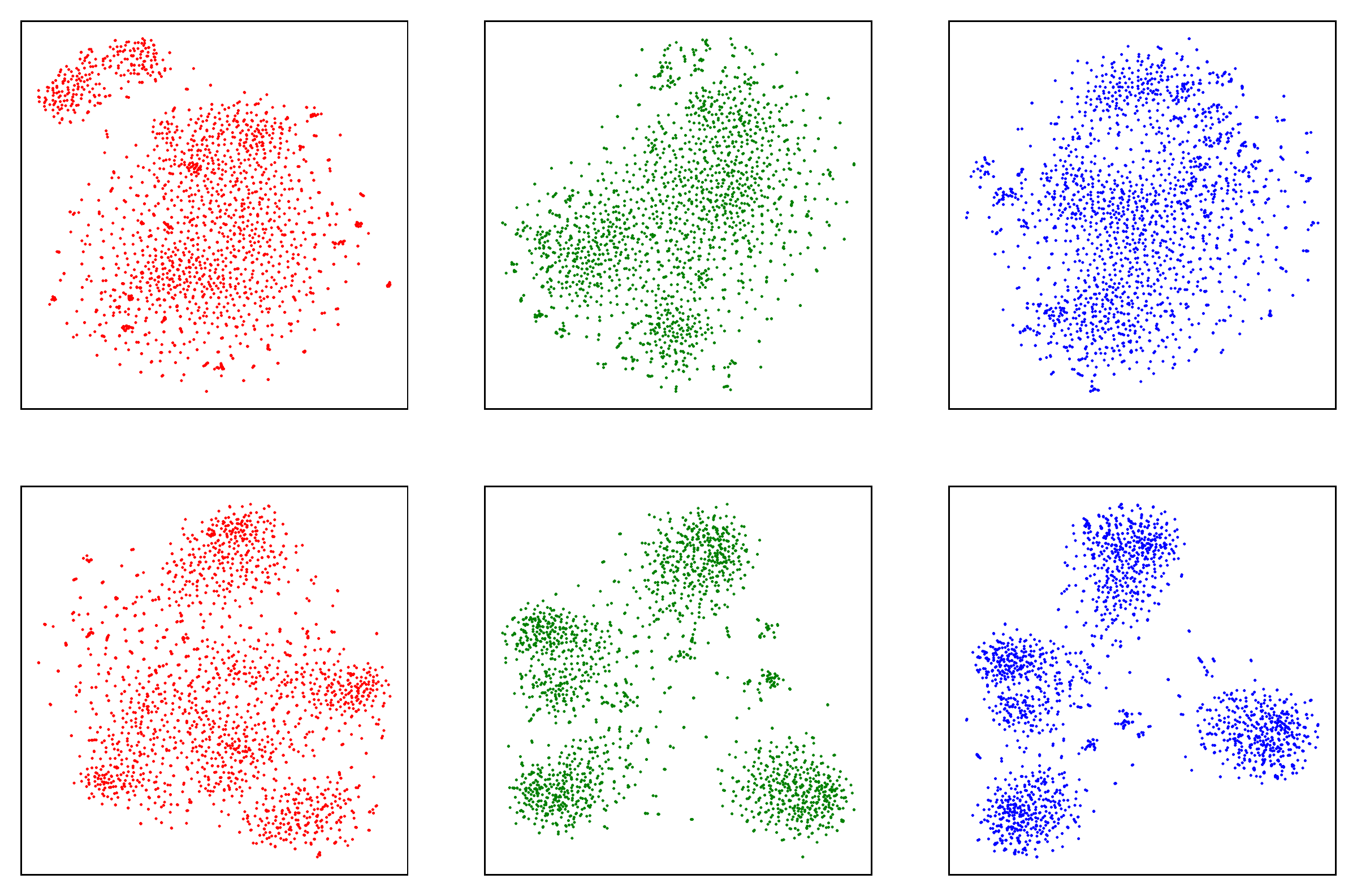}
    \caption{t-SNE results on decoder hidden states. Upper row: \base; lower row: \ours; column $k$ shows hidden states sampled from tokens at $k$ steps following a delimiter.}
    \label{fig:tsne}
\end{figure}

First, we simply calculate the average \textit{unique} predicted phrases produced by both \base and \ours in experiments shown in Section~\ref{subsec:scientific} (beam size is 50). 
The resulting numbers are 20.38 and 89.70 for \base and \ours respectively.
Second, from the model running on the \kpk validation set, we randomly sample 2000 decoder hidden states at $k$ steps following a delimiter ($k=1,2,3$) and apply an unsupervised clustering method (t-SNE \citep{tsne}) on them. From the Figure~\ref{fig:tsne} we can see that hidden states sampled from \ours are easier to cluster while hidden states sampled from \base yield one mass of vectors with no obvious distinct clusters. Results on both metrics suggest target encoding and orthogonal regularization indeed help diversifying generation of our model.


\subsection{Qualitative Analysis}

To illustrate the difference of predictions between our proposed models, we show an example chosen from the \kpk validation set in Appendix~\ref{appd:cherry_pick}. In this example there are 29 ground truth phrases. Neither of the models is able to generate all of the keyphrases, but it is obvious that the predictions from \base all start with ``test'', while predictions from \ours are diverse. This to some extent verifies our assumption that without the target encoder and orthogonal regularization, decoder states following delimiters are less diverse.


\section{Conclusion and Future Work}
We propose a recurrent generative model that sequentially generates multiple keyphrases, with two extra modules that enhance generation diversity.
We propose new metrics to evaluate keyphrase generation.
Our model shows competitive performance on a set of keyphrase generation datasets, including one introduced in this work.
In future work, we plan to investigate how target phrase order affects the generation behavior, and further explore set generation in an order invariant fashion.

\section*{Acknowledgments}
This work is supported by the National Science Foundation under grant No. 1525186. This research was also supported in part by the University of Pittsburgh Center for Research Computing through the resources provided. The authors thank the anonymous ACL reviewers for their helpful feedback and suggestions.

\bibliography{biblio}{}
\bibliographystyle{apalike}

\clearpage
\appendix

\section{Sequence to Sequence Generation}
\label{appd:basic_model}

\subsection{The Encoder-Decoder Model}
Given a source text consisting of $N$ words $w_e^1,\dots,w_e^N$, the encoder converts
their corresponding embeddings $x_e^1,\dots,x_e^N$ into a set of $N$ real-valued vectors $h_e=(h_e^1,\dots,h_e^{N})$ with a bidirectional GRU \citep{cho2014gru}: 
\begin{equation}
\begin{aligned}
h_{e,\mathrm{fwd}}^t&=\GRU_{e,\mathrm{fwd}}(x_e^t,h_{e,\mathrm{fwd}}^{t-1}),\\
h_{e,\mathrm{bwd}}^t&=\GRU_{e,\mathrm{bwd}}(x_e^t,h_{e,\mathrm{bwd}}^{t+1}),\\
h_e^t&=\langle h_{e,\mathrm{fwd}}^t,h_{e,\mathrm{bwd}}^t\rangle.
\end{aligned}
\end{equation}
Dropout \citep{srivastava2014dropout} is applied to both $x_e$ and $h_e$ for regularization.



The decoder is a uni-directional GRU, which generates a new state
$h_d^t$ at each time-step $t$ from the word embedding $x_d^{t}$ and the recurrent state $h_d^{t-1}$:
\begin{equation}
h_d^t=\GRU_d(x_d^{t},h_d^{t-1}).\footnote{During training (with teacher forcing), $w_d^{t}$ is the ground truth target token at previous time-step $t-1$; during evaluation, $w_d^{t}=y^{t-1}$, is the prediction at the previous time-step.}
\end{equation}

The initial state $h_d^0$ is derived from the final encoder state $h_e^{N}$
by applying a single-layer feed-forward neural net (FNN): 
\begin{equation}
h_d^0=L_\lti^{\tanh}(h_e^{N}). 
\end{equation}
Dropout is applied to both the embeddings $x_d$ and the GRU states $h_d$.

\subsection{Attentive Decoding}

When generating token $y^t$, in order to better incorporate information from the source text, an attention
mechanism \citep{bahdanau2014attntion} is employed to infer the importance $\alpha^{t,i}$ of each source word $w_e^i$
given the current decoder state $h_d^t$. This importance is measured by an energy function with a 2-layer FNN:
\begin{equation}
\mathrm{energy}(h_d^t,h_e^i)=L_\rlti{L:squash}(L_\lti^{\tanh}(\langle h_d^t,h_e^i\rangle)).
\end{equation}
The output over all decoding steps $t$ thus define a distribution over the source
sequence:
\begin{equation}
\label{eqn:alpha}
\alpha^t=\smx(\mathrm{energy}(h_d^t,h_e)).
\end{equation}

These attention scores are then used as weights for a refined representation of the source encodings, which is then concatenated to the decoder state $h_d^t$ to derive a generative distribution $p_a$:
\begin{equation}
\label{eqn:p_a}
p_a(y^t)=L_\lti^\smx(L_\lti^{\tanh}(\langle h_d^t,\sum_i\alpha^{t,i}\cdot h_e^i\rangle)),
\end{equation}
where the output size of $L_3$ equals to the target vocabulary size. Subscript $a$ indicates the \textit{abstractive} nature of $p_a$ since it is a distribution over a prescribed vocabulary.

\subsection{Pointer Softmax}
\label{sec-ptr-smx}
We employ the pointer softmax \citep{gulcehre2016pointersoftmax} mechanism to switch between generating a token $y^t$ (from a vocabulary) and pointing (to a token in the source text).
Specifically, the pointer softmax module computes a scalar switch $s^t$ at each generation time-step and uses it
to interpolate the abstractive distribution $p_a(y^t)$ over the vocabulary (see Equation~\ref{eqn:p_a}) and the extractive distribution $p_x(y^t)=\alpha^t$ over the source text tokens:
\begin{equation}
\label{eqn:p_yt}
p(y^t)=s^t\cdot p_a(y^t) + (1 - s^t) \cdot p_x(y^t),
\end{equation}
where $s^t$ is conditioned on both the attention-weighted source representation $\sum_i\alpha^{t,i}\cdot h_e^i$ and the decoder state $h_d^t$:
\begin{equation}
s^t=L_\lti^\sigm(\tanh(L_\lti(\sum_i\alpha^{t,i}\cdot h_e^i)+L_\lti(h_d^t))).
\end{equation}

\section{Experiment Results on \kpk Absent Subset}
\label{appd:kp20k_absent}

Generating \emph{absent} keyphrases on scientific publication datasets is a rather challenging problem.
Existing studies often achieve seemingly good performance by measuring recall on tens and sometimes hundreds of keyphrases produced by exhaustive decoding with a large beam size --- thus completely ignoring precision.


We report the models' Recall@10/50 scores on the \emph{absent} portion of five scientific paper datasets in Table~\ref{tab:absent} to be in line with previous studies. 


\begin{table*}[ht!]
  \centering
  \scriptsize
  \begin{tabular}{c|cc|cc|cc|cc|cc}
    \toprule
    \midrule
    & \multicolumn{2}{c|}{\textbf{Kp20K}}
    & \multicolumn{2}{c|}{\textbf{Inspec}}
    & \multicolumn{2}{c|}{\textbf{Krapivin}}
    & \multicolumn{2}{c|}{\textbf{NUS}}
    & \multicolumn{2}{c}{\textbf{SemEval}}
    \\
    \midrule
    Model & \textbf{R@10}& \textbf{R@50} & \textbf{R@10}& \textbf{R@50} & \textbf{R@10}& \textbf{R@50} & \textbf{R@10} & \textbf{R@50} & \textbf{R@10}& \textbf{R@50} \\  
    \midrule
    \midrule
    \text{CopyRNN}~\cite{meng2017deep} & \underline{11.5} & \textbf{18.9} & \underline{5.1} & \textbf{10.1} & \underline{11.6} & \textbf{19.5} & \underline{7.8} & \textbf{14.4} & \textbf{4.9} & \textbf{7.5}  \\  
    \text{CopyRNN*}~\cite{meng2017deep} & 3.3 & 8.7 & 4.0 & 8.3 & 4.0 & 8.1 & 2.4 & 8.1 & 0.5 & 2.6  \\  
    \midrule
    \midrule
    \text{\base}~(ours) & 6.0 & 6.2 & 2.8 & 2.9 & 7.0 & 7.4 & 3.7 & 3.1 & 2.5 & 2.5  \\  
    \text{\ours}~(ours) & \textbf{11.7} & \underline{15.1} & \textbf{5.2} & \underline{7.1} & \textbf{12.0} & \underline{14.5} & \textbf{8.4} & \underline{11.0} & \underline{4.6} & \underline{6.3}  \\  
    \midrule
    \bottomrule
  \end{tabular}
  \caption{Performance of absent keyphrase prediction on scientific publications datasets. Best/second-best performing score in each column is highlighted with bold/underline.}
  \label{tab:absent}
\end{table*}

The absent keyphrase prediction highly prefers recall-oriented models, therefore CopyRNN with beam size of 200 is innately proper for this task setting. However, from the results we observe that with the help of exhaustive decoding and diverse mechanisms, \ours is able to perform comparably to CopyRNN model, and it generally works better for top predictions. Even though the trend of models' performance somewhat matches what we observe on the \textit{present} data, we argue that it is hard to compare different models' performance on such scale. We argue that \stackex is better testbeds for \textit{absent} keyphrase generation.

\section{\stackex Data Collection}
\label{appd:stackex_collection}
We download the public data dump from \url{https://archive.org/details/stackexchange}, and choose 19 computer science related topics from Oct. 2017 dump.
We select computer science forums (CS/AI), using ``title'' + ``body'' as source text and ``tags'' as the target keyphrases. 
After removing questions without valid tags, we collect 330,965 questions. We thus randomly select 16,000 for validation, and another 16,000 as test set. 
Note some questions in StackExchange forums contain large blocks of code, resulting in long texts (sometimes more than 10,000 tokens after tokenization), this is difficult for most neural models to handle. 
Consequently, we truncate texts to 300 tokens and 1,000 tokens for training and evaluation splits respectively.

\section{Implementation Details}
\label{appd:implement_detail}

Implementation details of our proposed models are as follows.
In all experiments, the word embeddings are initialized with 100-dimensional random matrices.
The number of hidden units in both the encoder and decoder GRU are 150. 
The number of hidden units in target encoder GRU is 150. The size of vocabulary is 50,000.
In all experiments, we use a dropout rate of 0.1.

The numbers of hidden units in MLPs described in Section~\ref{sec:model} are as follows.
During negative sampling, we randomly sample 16 samples from the same batch, thus target encoding loss in Equation~\ref{eqn:target_encoder_loss} is a 17-way classification loss.
In \ours, we select both $\lambda_\mathrm{OR}$ and $\lambda_\mathrm{SC}$ in Equation~\ref{eqn:entire_loss} from [0.01, 0.03, 0.1, 0.3, 1.0] using validation sets.
The selected values are listed in Table~\ref{tab:lambdas}.

\begin{table}[h!]
    \centering
    \small
    \begin{tabular}{r|c|c|c}
        \toprule
        \midrule
        \multicolumn{2}{c|}{Experiment Setting} & $\lambda_\mathrm{OR}$ & $\lambda_\mathrm{SC}$  \\
        \midrule
        \midrule
        \multicolumn{2}{c|}{Table~\ref{tab:transfer}} & 1.0 & 0.03 \\
        \midrule
        \multicolumn{2}{c|}{Table~\ref{tab:stack_exchange}} & 0.03 & 0.1 \\
        \midrule
        Table~\ref{tab:f1_at_m}, \kpk & Greedy  & 1.0 & 0.3 \\
        \midrule
        Table~\ref{tab:f1_at_m}, \kpk & Top Rank  & 1.0 & 0.3 \\
        \midrule
        Table~\ref{tab:f1_at_m}, \stackex & Greedy  & 1.0 & 0.3 \\
        \midrule
        Table~\ref{tab:f1_at_m}, \stackex & Top Rank  & 1.0 & 0.3 \\
        \midrule
        \multicolumn{2}{c|}{Table~\ref{tab:ablation}, \base + Orth. Reg.}  & 0.3 & 0.0 \\
        \midrule
        \multicolumn{2}{c|}{Table~\ref{tab:ablation}, \base + Sem. Cov.}  & 0.0 & 0.03 \\
        \midrule
        \multicolumn{2}{c|}{Table~\ref{tab:ablation}, \ours}  & \multicolumn{2}{c}{Same as Table~\ref{tab:transfer}} \\
        \midrule
        \multicolumn{2}{c|}{Table~\ref{tab:absent}}  & \multicolumn{2}{c}{Same as Table~\ref{tab:transfer}} \\
        \midrule
        \bottomrule
        
    \end{tabular}
    \caption{Semantic coverage and orthogonal regularization coefficients.}
    \label{tab:lambdas}
\end{table}

We use \emph{Adam} \citep{kingma2014adam} as the step rule for optimization. The learning rate is $1e^{-3}$. The model is implemented using \textit{PyTorch}~\citep{paszke2017automatic} and OpenNMT~\citep{opennmt}.

For exhaustive decoding, we use a beam size of 50 and a maximum sequence length of 40.

Following \citet{meng2017deep}, lowercase and stemming are performed on both the ground truth and generated keyphrases during evaluation.

We leave out 2,000 data examples as validation set for both \kpk and \stackex and use them to identify optimal checkpoints for testing. And all the scores reported in this paper are from checkpoints with best performances (\fo) on validation set. 

In Section~\ref{subsection:tsne}, we use the default parameters for t-SNE in sklearn (learning rate is 200.0, number of iterations is 1000, as defined in \footnote{\url{https://scikit-learn.org/stable/modules/generated/sklearn.manifold.TSNE.html}}).



\section{Dataset and Evaluation Details}
\label{appd:evaluate_detail}
We strictly follow the data pre-processing and evaluation protocols provided by \citet{meng2017deep}.

We pre-process both document texts and ground-truth keyphrases, including word segmentation, lowercasing and replacing all digits with symbol $<$digit$>$. 
In the datasets, examples with empty ground-truth keyphrases are removed.

We evaluate models' performance on predicting present and absent phrases separately. 
Specifically, we first lowercase the text, then we determine the presence of each ground-truth keyphrase by checking whether it is a sub-string of the source text (we use Porter Stemmer \footnote{\url{https://www.nltk.org/api/nltk.stem.html\#module-nltk.stem.porter}}). 
To evaluate present phrase performance, we compute Precision/Recall/F1-score (see \ref{eq:precision}-\ref{eq:fscore} for formulas) for each document taking only present ground-truth keyphrases as target and ignore the absent ones. 

\begin{alignat*}{2}
    & P@k &&= \frac{\#(correct@k)}{\textit{min}\{k, \#(pred)\}} \numberthis \label{eq:precision} \\
    & R &&= \frac{\#(correct@k)}{\#(target)} \numberthis \\
    & F_1@k &&= \frac{2 * P@k * R}{P@k + R} \numberthis \label{eq:fscore}
\end{alignat*}

where $\#(pred)$ and $\#(target)$ are the number of predicted and ground-truth keyphrases respectively; and $\#(correct@k)$ is the number of correct predictions among the first $k$ results.

We report the macro-averaged scores over documents that have at least one present ground-truth phrases (corresponding to the column \#PreDoc in Table \ref{tab:dataset_stat}), and similarly to the case for absent phrase evaluation.

\begin{table}[t!]
    \centering
    \scriptsize
    \fontsize{6}{8}\selectfont
    \begin{tabular}{l|r|r|r|r|r|r}
        \toprule
        \midrule
        Dataset & \#Doc & \#KP & \#PreDoc & \#PreKP & \#AbsDoc & \#AbsKP\\
        \midrule
        \midrule
        \text{\kpk} & 19,987 & 105,181 & 19,048 & 66,595 & 16,357 & 38,586  \\
        \text{\inspec} & 500 & 4,913 & 497 & 3,858 & 381 & 1,055 \\
        \text{\krapivin} & 460 & 2,641 & 437 & 1,485 & 417 & 1,156 \\
        \text{\nus} & 211 & 2,461 & 207 & 1,263 & 195 & 1,198 \\
        \text{\semeval} & 100 & 1,507 & 100 & 671 & 99 & 836\\
        \text{\stackex} & 16,000 & 43,131 & 13,475 & 24,809 & 10,984 & 18,322 \\
        \text{\duc} & 308 & 2,484 & 308 & 2,421 & 38 & 63 \\
        \midrule
        \bottomrule
    \end{tabular}
    \caption{Statistics on number of documents and keyphrases of each test set. \#Doc\#KP denotes the number of documents/ground-truth keyphrases in the dataset. \#PreKP/\#AbsKP denotes the number of present/absent ground-truth keyphrases, and \#PreDoc/\#AbsDoc denotes the number of documents that contain at least one present/absent ground-truth keyphrase.}
    \label{tab:dataset_stat}
\end{table}

\section{Examples of \kpk and \stackex with Model Prediction}
\label{appd:cherry_pick}
\begin{figure*}[t]
    \centering
    \includegraphics[width=1.0\textwidth]{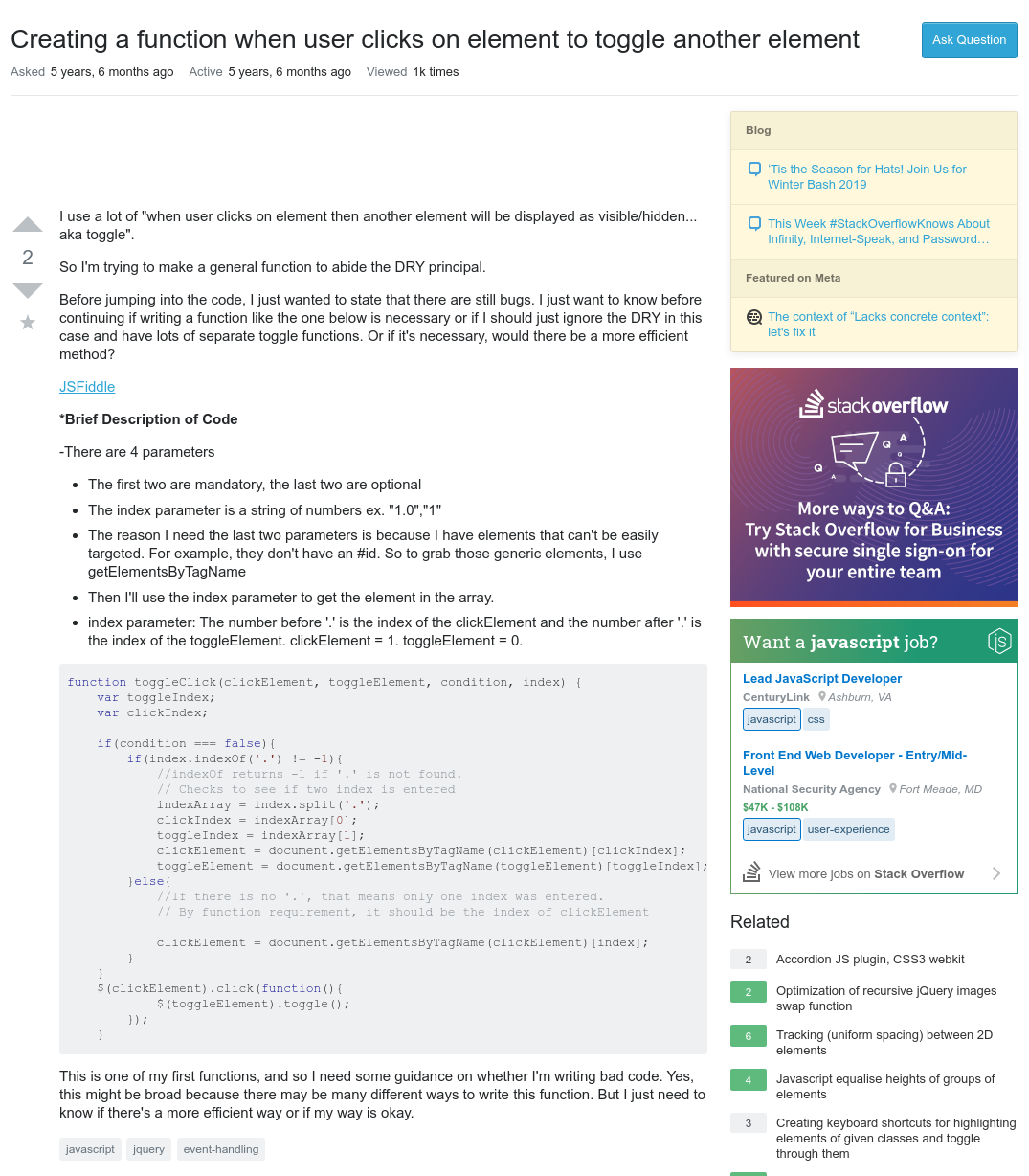}
    \caption{Example from the \stackex dataset, we show the screenshot of the original web page to better present the example. Note the input to the model is the entire question (including the code), we removed the format information in the dataset. Also note on the bottom of the screenshot it shows the 3 keyphrases (in this example all absent) which we collected as the ground-truth keyphrases in our dataset.\\
    \textbf{Ground Truth:} \textcolor{red}{javascript} ; \textcolor{red}{jquery} ; \textcolor{red}{event handling} \\
    \textbf{\base Prediction:} \textcolor{red}{javascript}; c$\#$; \textcolor{red}{jquery}; php; linq; comparative review; ecmascript 6; asp . js; beginner; strings; performance; datetime \\
    \textbf{\ours Prediction:} \textcolor{red}{javascript} ; \textcolor{red}{jquery} ; performance ; \textcolor{red}{event handling} ; array ; twitter bootstrap ; beginner ; algorithm ; indexarray ; optimization ; event programming ; datetime ; comparative review ; ecmascript 6 ; indexof ; dry ; php ; r ; java ; coffeescript ; combinatorics ; dom ; html ; event tracking ; strings ; python ; ruby ; natural language processing ; animation ; angular . js ; homework ; parameters ; jquery ui ; functional programming ; google app engine ; . net ; python 2 . 7 ; c$\#$ ; php5 ; validation ; regex ; parsing ; formatting ; hash table ; object oriented ; web scraping ; python 3 . x ; python 3 . x programming ; python 2 . net ; python 2 . 6 ; python 2 . sql ; mysql ; object oriented design ; actionscript}
    \label{fig:stackex_cherry}
\end{figure*}

See Table~\ref{tab:cherry_pick} and Figure~\ref{fig:stackex_cherry}.

\begin{table*}[h!]
    \centering
    \footnotesize
    \begin{tabular}{r|l}
        \toprule
        Source & Integration of a \colorbox{blue!30}{Voice Recognition} System in a \colorbox{blue!30}{Social Robot}  \\
               & \colorbox{blue!30}{Human-robot interaction} (\colorbox{blue!30}{HRI}) (1) is one of the main fields in the study and research of robotics. \\
               & Within this field, dialogue systems and interaction by voice play an important role. When speaking about \\
               & human-robot natural \colorbox{blue!30}{dialogue} we assume that the robot has the capability to accurately recognize what the \\
               & human wants to transmit verbally and even its semantic meaning, but this is not always achieved.\\
               & In this article we describe the steps and requirements that we went through in order to endow \\
               &  the personal social robot \colorbox{blue!30}{Maggie}, developed at the University Carlos III of Madrid, with the capability of  \\
               & understanding the natural language spoken by any human. We have analyzed the \\
               & different possibilities offered by current software/hardware alternatives by testing them in real environments.\\
               & We have obtained accurate data related to the \colorbox{blue!30}{speech recognition} capabilities in different environments, \\
               & using the most modern audio acquisition systems and analyzing not so typical parameters \\
               & such as user age, gender, intonation, volume, and language. \\
               & Finally, we propose a new model to classify recognition results as accepted or rejected, \\
               & based on a second \colorbox{blue!30}{automatic speech recognition} (\colorbox{blue!30}{ASR}) opinion.This new approach takes into account \\
               &  the precalculated success rate in noise intervals for each recognition framework, decreasing the rate of \\
               & false positives and false negatives. \\
        \midrule
        \base  &  voice recognition system ;  \textbf{social robot} ; \textbf{human robot interaction} ; \textbf{voice recognition} ; hri ; \textbf{speech recognition} ; \\
        &  \textbf{automatic speech recognition} ; noise intervals ; noise ; human robot ; automatic speech ; natural language \\
        \midrule
        \ours  &  \textbf{human robot interaction} ; \textbf{voice recognition} ; \textbf{social robotics} ; \textbf{social robots} ; integration ; \textbf{speech recognition} ; \\
        &   hri ; \textbf{social robot} ; robotics ; voice recognition system ; recognition ; \textbf{asr} ; \textbf{automatic speech recognition} ;  \\
        \midrule
 Ground Truth  &  asr ; automatic speech recognition ; dialogue ;  human robot interaction ; maggie ; social robot ; \\
               & speech recognition ; voice recognition ;\\
        \bottomrule
    \end{tabular}
    \caption{Example from \kpk validation set, and predictions generated by \base and \ours models.}
    \label{tab:cherry_pick}.
\end{table*}

\clearpage

\end{document}